\documentclass[letterpaper]{article} 
\usepackage[preprint]{aaai2027}  
\usepackage[hyphens]{url}  
\usepackage{graphicx} 
\usepackage{natbib}  
\usepackage{caption} 
\usepackage{algorithm}
\usepackage{algorithmic}

\usepackage{newfloat}
\usepackage{listings}
\DeclareCaptionStyle{ruled}{labelfont=normalfont,labelsep=colon,strut=off} 
\floatstyle{ruled}
\newfloat{listing}{tb}{lst}{}
\floatname{listing}{Listing}

\usepackage{booktabs}
\usepackage{amsmath}
\usepackage{amssymb}
\usepackage{dsfont}
\usepackage{subcaption}

\title{Self-Keyed In-Parameter Memorization for External Memory-Free\\Dialogue Recall in Frozen Language Models}
\author {
    XinYu Piao,
    Myunghoon Ryu,
    Joongheon Kim\corresponding,
    Jong-Kook Kim\corresponding
}
\affiliations {
    Department of Electrical and Computer Engineering, Korea University, Seoul, Republic of Korea\\
    \{xypiao97, ryumh10, joongheon, jongkook\}@korea.ac.kr
}

\begin{document}

\maketitle

\begin{abstract}

A large language model maintains conversational content in an external storage that is outside the model and grows as the dialogue continues.
This work assumes an environment where a language model has no external memory and no prior knowledge of the future question or its answer. 
The motivation is to research whether a dialogue can be stored in the model parameters and later answer a question about it. 
This work proposes self-keyed in-parameter memorization (\textsc{Skim}), a closed-form method in which the model generates its own questions about each dialogue turn and writes those questions into its parameters as the keys that later retrieve the turn's content. 
Generating many varied questions for each turn allows a later user question to match one even when phrased differently, and its answer is recalled from the parameters alone.
Evaluated on five language models, \textsc{Skim} recalls most stored dialogue turns from later user questions,  and it answers from the model parameters without any external storage.

\end{abstract}

\section{Introduction}
Recent large language models (LLMs) serve as conversational assistants and their conversations often run for many exchanges~\cite{petroni2019language,roberts2020much,de2021editing,kwon2023efficientmemorymanagementlarge}. 
As a conversation lengthens, answering well often depends on recalling something the user mentioned much earlier~\cite{xu2022beyond,packer2023memgpt,wu2025longmemevalbenchmarkingchatassistants}. 
However, language models keep the dialogue not in their parameters but in a key-value (KV) cache or an external retrieval store that grows as the history accumulates. 
The KV cache stores the attention keys and values of every past token, sparing the model from recomputing them at each step~\cite{vaswani2017attention,shazeer2019fasttransformerdecodingwritehead,kwon2023efficientmemorymanagementlarge}. 
Its size grows proportionately to the length of the history and the generated tokens accumulate in the KV cache. 
Both memory use and inference cost rise as more turns arrive~\cite{pope2022efficiently,dao2022flashattentionfastmemoryefficientexact}. 
Instead, retrieval-augmented generation (RAG) stores the dialogue in an external index and searches it for relevant passages at each question, placing them in the context~\cite{lewis2020retrieval,lewis2021retrievalaugmentedgenerationknowledgeintensivenlp}. 
The answer then depends on whether the index holds the right passage and whether the question was turned into an effective search query~\cite{guu2020realmretrievalaugmentedlanguagemodel,maharana2024evaluatinglongtermconversationalmemory}.
Based on these observations, this work asks whether a frozen model can internalize a dialogue as it unfolds and later answer a question about it from its parameters alone.
The aim is to establish whether such internalization is feasible at all, rather than to improve on retrieval or long-context systems that already answer from external memory.

Prior studies aim to store a model's memory in its parameters rather than in an external buffer~\cite{meng2022locating,meng2022mass,wang2024memoryllmselfupdatablelargelanguage,fang2025alphaedit}. 
Weight editing is one such approach that writes a fact directly into the parameters~\cite{mitchell2022fastmodeleditingscale, meng2022locating, meng2022mass}. 
Given a subject and a relation, such as a person and their birthplace, these methods compute a closed-form update to one or a few feed-forward projections that causes the model to output the intended object~\cite{meng2022locating, meng2022mass}.
A null-space projection allows many such edits to accumulate without overwriting one another~\cite{fang2025alphaedit}. 
However, this approach rests on two assumptions that conversational memory does not satisfy. 
First, the object to store must be provided as a label because the update is solved to reproduce that object at the edited site~\cite{meng2022locating,meng2022mass}. 
Because no such label exists in a dialogue, the model must itself determine which part of a raw user turn should be stored.
Second, a fixed cue that is typically the subject and relation must be known in advance when constructing the edit~\cite{meng2022locating,meng2022mass}.
In a dialogue, there is no fixed cue to install because the later question is unknown when the turn is stored and may be phrased in unforeseen ways. 
Long-context models and retrieval systems avoid both assumptions, but they retain the dialogue outside the parameters~\cite{lewis2021retrievalaugmentedgenerationknowledgeintensivenlp}, in a cache or an index that grows as turns accumulate~\cite{maharana2024evaluatinglongtermconversationalmemory,kwon2023efficientmemorymanagementlarge}. 
Thus, editing requires supervision that a dialogue cannot provide, whereas external buffers avoid that supervision only by incurring an ever-growing storage cost~\cite{kwon2023efficientmemorymanagementlarge, maharana2024evaluatinglongtermconversationalmemory}.

Prior work established that the feed-forward layers of a transformer store factual associations and recall them at inference, and that a band of middle layers is where such facts are most readily located and edited \cite{geva2021transformerfeedforwardlayerskeyvalue,dai2022knowledgeneuronspretrainedtransformers,meng2022locating,geva2023dissecting}. 
Building on that finding, this work writes each user turn into the down-projection of one mid-to-late feed-forward layer by a closed-form rule. 
The target layer is chosen per model and a later question is answered from the edited parameters alone. 
This work assumes an environment where a frozen model has no external memory and no access to the future question or its answer when it stores a turn, which makes a direct comparison to other editing methods difficult. 
The setting imposes four constraints, namely no gradient fine-tuning, no external memory at recall, a per-turn cost independent of the number of stored turns, and preserved general behavior.
This work forgoes the KV cache despite its usefulness and keeps the dialogue in the parameters~\cite{vaswani2017attention,shazeer2019fasttransformerdecodingwritehead}.
Because the model has no future question or answer, it must produce from the turn alone the object and cue that weight editing receives but a dialogue cannot provide.

This work proposes a Self-Keyed In-parameter Memorization approach, called \textsc{Skim}, that allows a frozen model to store a turn and later recall it without the ground-truth answer or the future question. 
The proposed \textsc{Skim} operates on spans, each a short contiguous piece of a turn's text. 
\textsc{Skim} proceeds in three stages, namely (i) extracting candidate spans from the raw turn, (ii) generating for each span several questions that the span answers and keeping the ones that best separate that span from its rivals in the same turn, and (iii) installing every question--span pair into one projection by a joint closed-form edit. 
The generated questions take the place of the fixed subject--relation input that prior editing must be given in advance. Because many diverse questions serve as keys for each span, a later user question resembles one of them even when phrased differently, and the question itself selects the right span at recall. 
The main contributions of this work are summarized as follows.
\begin{itemize}
\item This work aims to retain a dialogue and answer a later question about it in an environment without external memory, neither a KV cache nor a retrieval store.
\item This work proposes \textsc{Skim}, a closed-form method that internalizes each turn into the model parameters without the ground-truth answer or the future question, using self-generated questions as keys for its stored spans.
\item This work shows that \textsc{Skim} stores a dialogue as its turns arrive one at a time and reaches the recall of a single batch edit without the forgetting that sequential edits cause.
\item Evaluated on five language models, \textsc{Skim} recalls stored dialogue turns at a five-model mean of 63.1\% and 75.0\% on the strongest model, well above the no-cache floor that removes all memory.
\end{itemize}
\section{Problem Setting and Preliminaries}

\paragraph{Assumptions.} 
Current language models keep their parameters fixed and hold each conversation in an external memory that grows as the dialogue lengthens~\cite{lewis2021retrievalaugmentedgenerationknowledgeintensivenlp, maharana2024evaluatinglongtermconversationalmemory}. 
This paper asks whether a frozen model can hold the conversation in its parameters instead of an external memory, and answer later questions from them.
The assumed scenario is as follows.
A stream of user turns $u_1,\dots,u_N$ arrives, and each $u_i$ states a fact that some later question $q_i$ asks about, whose answer is $a_i$. 
The language model stores each turn in its parameters as it arrives, and it is more constrained than the external store it replaces.
A cache or a retrieval index maintains a turn in full and delays every decision until the query arrives. 
Instead, the model must determine from the turn itself what to store. 
It is given neither the later question $q_i$ that would identify the relevant content nor the answer $a_i$ that would specify the target. 
The phrasing of the later question may also be unanticipated.
When the later question arrives, the model receives it as the only input and must generate the answer. 
No session text, cache, or retrieval index remains available. 
Four constraints hold throughout. 
Storing a turn uses no gradient fine-tuning, its cost does not grow as more turns are stored, recall is a single forward pass, and the edit must leave the general behavior of the model intact.
\begin{figure*}[t]
    \vspace{0.4em}
    \centering
    \includegraphics[width=1\linewidth]{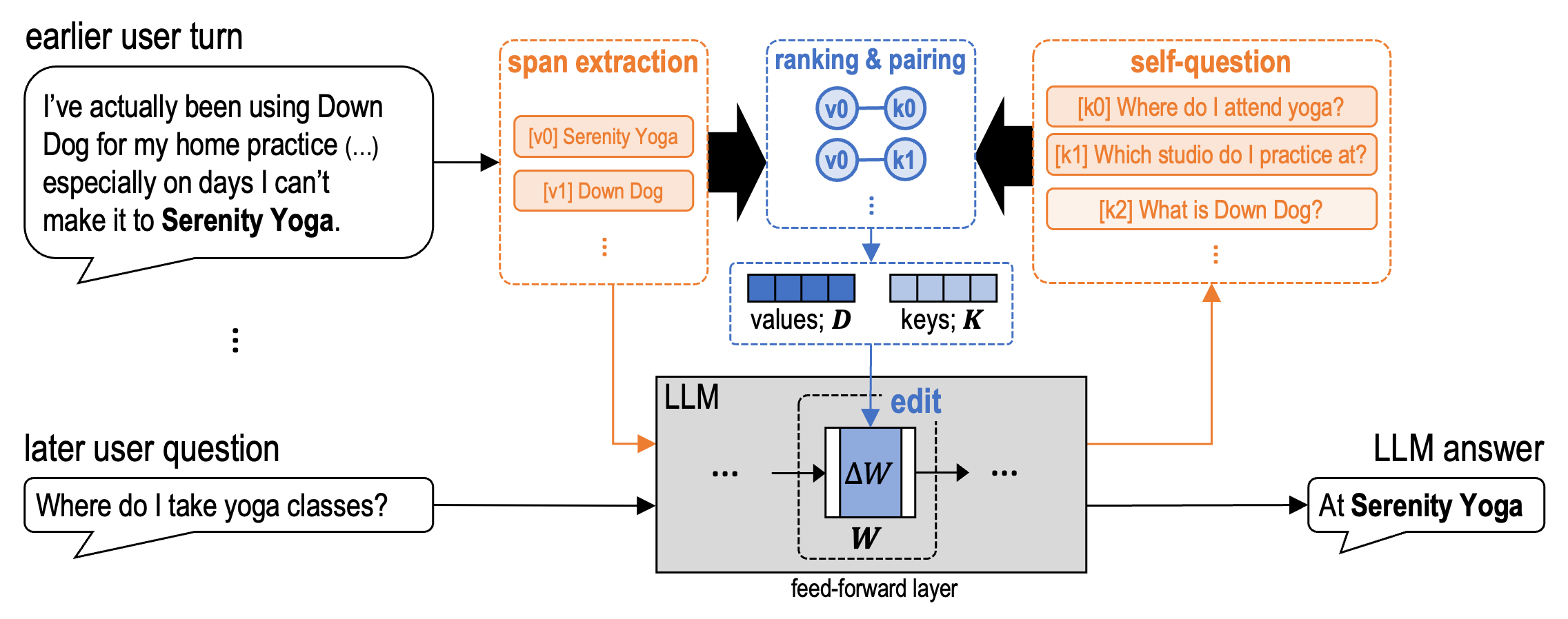}
    \vspace{0.2em}
    \caption{Overview of \textsc{Skim}. From an earlier turn the model extracts candidate spans as values and self-generates questions as keys. Both steps are answer-free, and a closed-form edit installs these pairs into one frozen feed-forward layer. A later question phrased differently forms a key that retrieves the value, and the model answers from its parameters.}
    \label{fig:overview}
\end{figure*}

\paragraph{Preliminaries.} 
The transformer feed-forward down-projection $W$ in a block $W\,\sigma(W_{\mathrm{up}}x)$ behaves as a key-value memory, where the hidden activation $k=\sigma(W_{\mathrm{up}}x)$ serves as an address and $Wk$ is the content read from it.
A single factual association is one entry of this memory, and editing $W$ overwrites the value retrieved by a given key reads\cite{geva2021transformerfeedforwardlayerskeyvalue,meng2022locating}.
To store $n$ associations, their keys form the columns of a matrix $K$ and their values the columns of a matrix $D$.
The edit $\Delta W$ that reproduces all of them minimizes a regularized least-squares objective,
\begin{align}
\|\Delta W K - D\|_F^2 + \lambda\|\Delta W\|_F^2,
\end{align}
where $\lambda>0$ sets the regularization strength.
Setting its gradient to zero gives the normal equations $\Delta W(KK^{\top}+\lambda I)=DK^{\top}$, whose closed-form ridge solution \cite{hoerl1970ridge} is
\begin{align}
\begin{split}
\Delta W & = DK^{\top}(KK^{\top}+\lambda I)^{-1} \\[3pt]
         & = D\,(K^{\top}K + \lambda I)^{-1}K^{\top}. \label{eq:ridge}
\end{split}
\end{align}
The second form inverts an $n\times n$ matrix, small when few turns are stored.
The edit reproduces each value $d_j$ at its key $k_j$ up to the regularizer.
A larger $\lambda$ keeps the edit small and favors preservation of the base weights, and a smaller one favors exact reproduction.
The update stores all $n$ values in superposition and recovers each as long as the keys remain near-orthogonal, which bounds the number of associations a layer holds before interference.
The solution requires no gradient descent and underlies the weight-editing family~\cite{meng2022locating,meng2022mass}.
The objective does not specify the origin of $K$ and $D$.
Prior editing methods are given a ground-truth value and a fixed key, whereas a dialogue provides neither of these. 
The method of the next section constructs both from the turn itself.

\section{Proposed Method}
As shown in Figure~\ref{fig:overview}, \textsc{Skim} extracts candidate spans as the values to store, generates questions about each span as the keys that address them, and installs the pairs into one feed-forward layer by the closed-form joint edit.
The first two stages use the frozen model only to generate and rank candidates, and only the third stage changes a parameter.
At recall, the model answers the user question from its parameters. 

\subsection{Self-Keyed Construction} 
The frozen language model produces short answer phrases when prompted to list the specific facts the turn $u$ states. 
\textsc{Skim} also takes the turn's quoted titles and capitalized entity phrases directly from its text. 
These candidates are filtered to short spans of a few words rather than the whole utterance and form the candidate value set $\mathcal{S}(u)$. 
The model then ranks each span $s\in\mathcal{S}(u)$ answer-free by
\begin{equation}
\rho(s) \;=\; \mathbb{E}_{q\sim Q_0(s)}\!\big[\log p_M(s \mid q, u)\big] \;+\; \beta\,\pi(s),
\label{eq:harvest}
\end{equation}
where $Q_0(s)$ is a set of self-questions the model generates for $s$, and $p_M(x\mid y)$ is the probability the model $M$ assigns to a span $x$ as the continuation of a prompt $y$. 
The first term rewards a span the model can recover from its own questions about it, which is what later makes the span addressable. 
The second term is the specificity prior $\pi(s)$ that equals the span's negative log-probability given a neutral prompt and rewards a rare span that a later question is more likely to ask about. 
The coefficient $\beta$ balances recoverability against specificity. 
When a turn is stored, the model does not yet know which span the later question will ask about, and the single highest-$\rho$ span is often not that one. 
Keeping top-$k$ spans as candidate values therefore defers the choice to recall time, where the later question retrieves the span that matches it.

Each span also needs keys that address it for a later question phrased differently at recall, and these keys are questions the model generates about the span itself.
For each span $s$, the model is prompted in $k_q$ fixed styles to generate a question whose answer is $s$, producing candidates $q\sim G(\cdot\mid s, u)$.
Each candidate is scored by a discriminative margin
\begin{align}
\begin{split}
\mu(q,s) = & \log p_M(s\mid q, u) \\[3pt]
&- \max_{s'\in \mathcal{S}(u)\setminus\{s\}} \log p_M(s'\mid q, u).
\end{split}
\label{eq:margin}
\end{align}
A high margin marks a question that points to $s$ and not to any rival, which makes the key discriminative and limits the chance that a question about one span retrieves another at recall.
Each span keeps its top $m$ questions by $\mu$ after rival surface forms are removed from their wording. 
The survivors are then expanded by deterministic string variants of the question text.
For the span "Serenity Yoga", a surviving question is "Which studio do I practice at?".
The question-word exchange swaps its "Which" for "What", and re-framing wraps it as "Earlier, the answer to `Which studio do I practice at?' was".
The resulting key set $R_s$ replaces the fixed subject--relation cue of prior editing, and its diversity lets a differently-phrased later question resemble one of the stored keys.

\subsection{Joint Edit in One Feed-Forward Layer}
Let $(r_j, s_j)$ enumerate the surviving question--span pairs of every turn accumulated to this point.
Each question $r_j$ is rendered in a fixed recall frame $\mathrm{frame}(\cdot)$ that presents it to the model as a prompt. 
The key $k_j$ is the hidden state that enters the down-projection of layer $\ell$ at the answer position~\cite{geva2021transformerfeedforwardlayerskeyvalue},
\begin{equation}
k_j \;=\; h^{(\ell)}\!\big(\mathrm{frame}(r_j)\big),
\label{eq:key}
\end{equation}
where $h^{(\ell)}(y)$ denotes that hidden state for a prompt $y$. 
Because this activation also enters $W$ at recall, the storage key equals the recall key in the same frame. 
The value $d_j$ is a residual output delta that makes the span the frame's continuation, scaled by the strength $f$,
\begin{align}
\begin{split}
d_j & = f\,\delta_j, \\[3pt]
\delta_j = \arg\min_{\delta} -&\log p_{M[\delta]}\!\big(s_j \mid \mathrm{frame}(r_j)\big),
\end{split}
\label{eq:value}
\end{align}
where $M[\delta]$ adds $\delta$ to the layer output at the answer position.  
The objective has no closed form because the span probability is nonlinear in $\delta$. 
A short gradient optimization finds $\delta$ while the weights stay frozen~\cite{meng2022locating}. 
The strength $f$ scales the stored value. 
A larger $f$ strengthens recall but pushes the residual stream farther from its base state. 
Recall and preserved behavior are therefore in tension. 
The keys $k_j$ are the columns of a matrix $K$ and the values $d_j$ are the columns of a matrix $D$. 
The closed-form solve of Eq.~\eqref{eq:ridge}~\cite{meng2022mass} fits all accumulated pairs jointly rather than one at a time and returns the single matrix $\Delta W$ that at once satisfies
\begin{equation}
\Delta W\, k_j \approx d_j \quad \text{for all } j .
\end{equation}
The ridge term $\lambda I$ keeps the inverse well conditioned when two keys are close and makes each match approximate rather than exact. 
Adding this $\Delta W$ to the down-projection $W$ of layer $\ell$ installs all pairs in one matrix. 
For an input $h$ the edited layer computes
\begin{equation}
(W+\Delta W)\, h = W h + \Delta W h .
\end{equation}
For an input unrelated to any stored key the extra term $\Delta W h$ stays small and the base behavior of the layer is preserved. 
A later question that forms a stored key $k_j$ makes the extra term contribute $d_j$ to the residual stream. 
The optimized $d_j$ raises the span probability and the model emits that span. 
The edit modifies a single mid-to-late down-projection that serves as the value memory of the feed-forward block.

\subsection{Recall and Self-Probed Selection}
At recall, the user question $q$ rendered in the frame that built the keys lands on a stored key and retrieves its value. 
The edited model $M_{\Delta W}$ then generates the answer $\hat{a}$ by decoding a short budget of tokens.
\begin{equation}
\hat{a} = \mathrm{decode}\big(M_{\Delta W},\, \mathrm{frame}(q)\big) .
\label{eq:readout}
\end{equation}
$M_{\Delta W}$ is the model after the edit $\Delta W$, and $\mathrm{frame}(q)$ is the user question in the recall frame. 
The decode preset is chosen per model and is either greedy decoding that suppresses repetition or a small beam that blocks repeated $n$-grams. 
It stops generation at an answer-completion criterion or a sentence-final token and makes the answer emerge cleanly.

The same pipeline runs for every model and only a few per-model settings vary. 
They form the configuration
\begin{equation}
\theta \;=\; (\ell,\ f,\ \text{decode preset},\ \text{key budget}) .
\end{equation}
These settings are fixed when the turn is stored and cannot be tuned to the later question. 
A self-probe selects them answer-free by standing in for the absent question. It probes each item by an unused paraphrase of the item's own self-generated question and records whether the stored span is emitted cleanly. 
Clean emission means the generation contains the span and avoids third-order token repetition and keeps lexical diversity above $0.5$. 
The clean fraction of the $n$ stored items is
\begin{equation}
\mathrm{af}_{\mathrm{clean}}(\theta) = \frac{1}{n}\sum_{i=1}^{n} \mathds{1}\big[\text{item } i \text{ emits its span cleanly}\big] .
\label{eq:afclean}
\end{equation}
This fraction needs no ground-truth answer or later question. The chosen configuration maximizes it.
\begin{equation}
\theta^{\star} \;=\; \arg\max_{\theta}\ \mathrm{af}_{\mathrm{clean}}(\theta) .
\end{equation}
The ranking is reliable because the probe repeats the recall operation itself on a stand-in and accesses no ground-truth answer. 
A configuration that degenerates or never retrieves the span fails the clean test and true recall alike. 
The clean fraction therefore tracks true recall and the method tunes each model from these self-probes alone.

\begin{figure*}[t]
\centering
\begin{subfigure}[t]{0.38\textwidth}
    \centering
    \includegraphics[width=\linewidth]{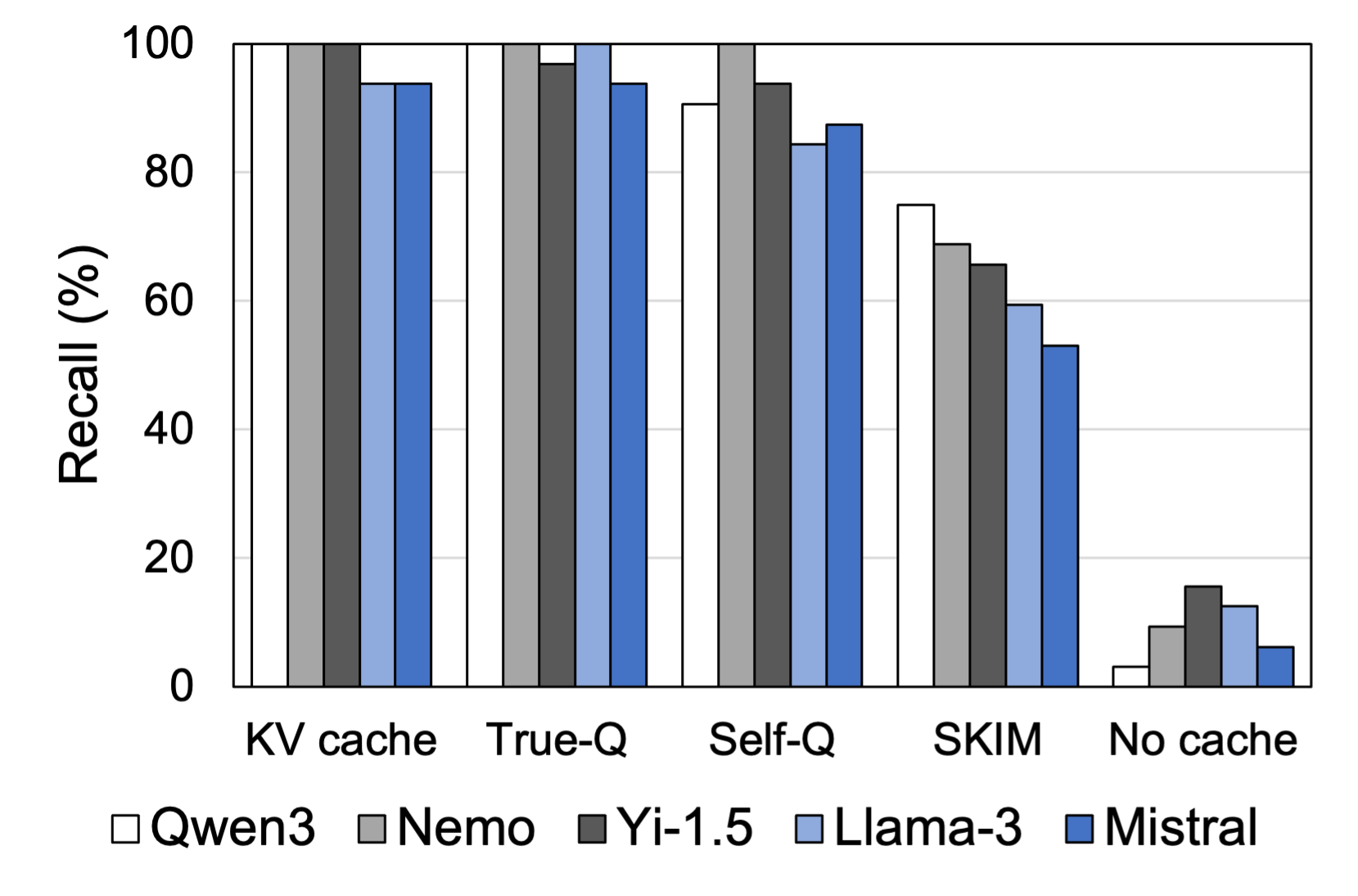}
    \caption{}
    \label{fig:headline}
\end{subfigure}
\hfill
\begin{subfigure}[t]{0.30\textwidth}
    \centering
    \includegraphics[width=\linewidth]{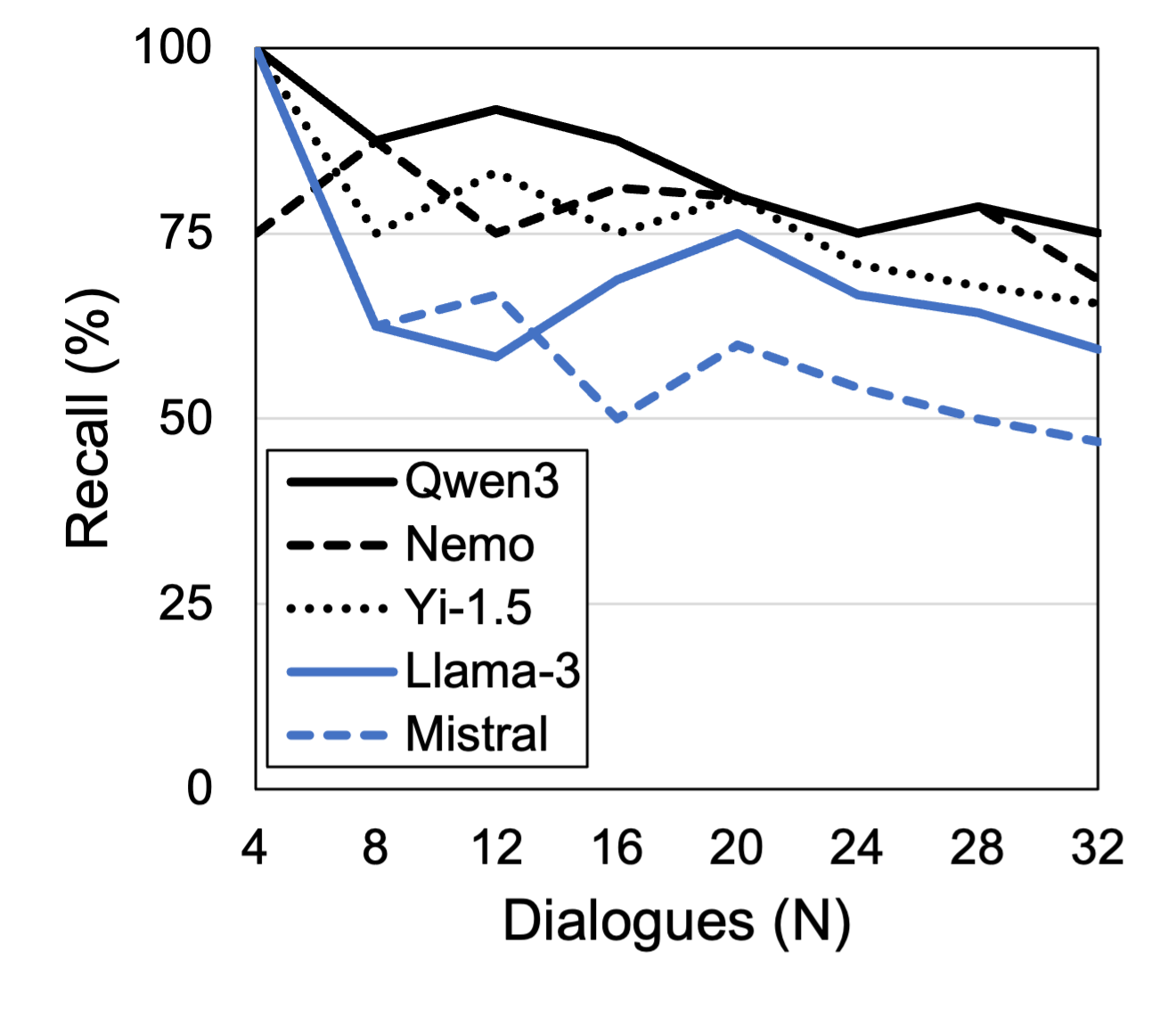}
    \caption{}
    \label{fig:n_scaling_recall}
\end{subfigure}
\hfill
\begin{subfigure}[t]{0.30\textwidth}
    \centering
    \includegraphics[width=\linewidth]{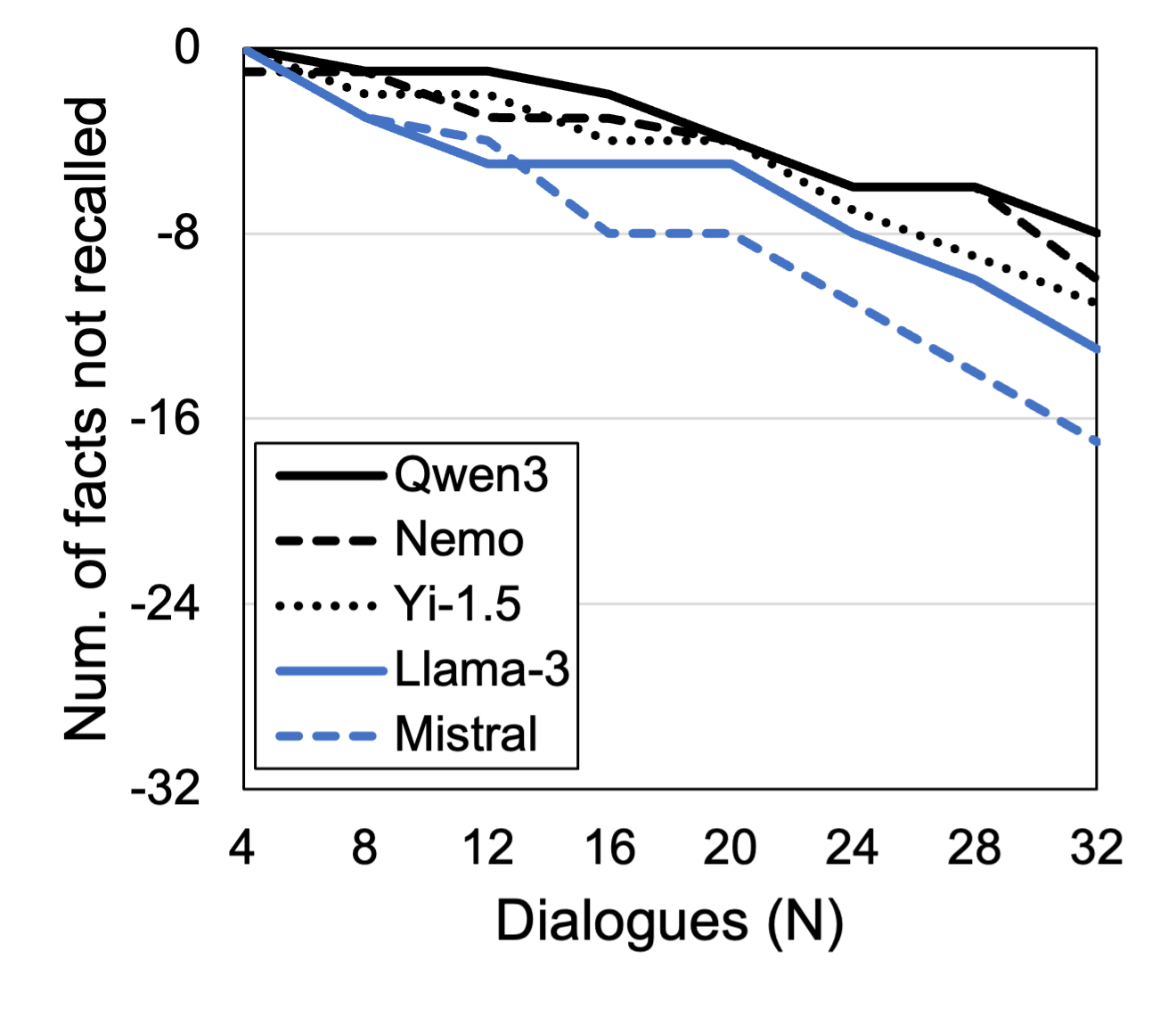}
    \caption{}
    \label{fig:n_scaling_fail}
\end{subfigure}
\caption{
Recall (\%) for five language models. Panel (a) reports recall at $N{=}32$ in five conditions from ceiling to floor, the KV cache upper bound, the answer-given references true-Q and self-Q, the answer-free SKIM, and a no-cache floor that removes all memory. Panels (b) and (c) trace recall and the count of unrecalled facts as $N$ grows from $4$ to $32$.
}
\label{fig:nscaling}
\end{figure*}

\subsection{Mitigating Forgetting from Sequential Edits}
A dialogue's turns arrive one at a time and each is written into the parameters as it arrives before any later turn is available. 
Such a memory is useful only if two demands are met. 
The first demand is that a new write leave the already stored turns intact because a growing sequence of edits can otherwise interfere and erase them. 
The second is that the memory the edit occupies stay fixed in size however long the dialogue runs which a growing cache cannot offer.
Meeting both while turns still arrive is not automatic because an edit that keeps growing as new turns come can disturb the turns already stored. 
The reference point is a batch edit that solves Eq.~\eqref{eq:ridge} once on the complete set of pairs~\cite{meng2022mass} and is the strongest result the closed-form rule can reach. 
\textsc{Skim} reproduces this same $\Delta W$ turn by turn even as turns still arrive. 
The turns arrive in chunks and the stored pairs accumulate into two growing matrices.
\begin{equation}
\begin{split}
K_{\le t}=[K_1,\dots,K_t], \\[3pt]
D_{\le t}=[D_1,\dots,D_t] .
\end{split}
\end{equation}
At each chunk, \textsc{Skim} discards the previous edit and re-solves from the original weights on every pair to this point.
\begin{equation}
\Delta W^{(t)} = D_{\le t}\big(K_{\le t}^{\top}K_{\le t}+\lambda I\big)^{-1}K_{\le t}^{\top} .
\label{eq:accum}
\end{equation}
This reads only $K_{\le t}$ and $D_{\le t}$ and any permutation $P$ of the columns leaves it unchanged because the two copies of $P$ cancel.
\begin{equation}
P^{\top}P = I .
\end{equation}
Arrival order and grouping therefore have no effect and the final edit equals the batch edit. 
A simpler additive scheme
\begin{equation}
\Delta W \leftarrow \Delta W + D_t(K_t^{\top}K_t+\lambda I)^{-1}K_t^{\top}
\end{equation}
never inverts the joint matrix $K_{\le t}^{\top}K_{\le t}$ and lets earlier turns drift~\cite{fang2025alphaedit} whereas re-solving from the original weights avoids the drift. 
The stored result is one fixed-size matrix $\Delta W$ and a question is a single forward pass and both stay constant however many turns are stored unlike a growing cache. 
The re-solve grows as turns accumulate but runs offline once per chunk and never enlarges the store or slows the answer.

\section{Evaluation and Analysis}

\subsection{Experimental Setup}
\label{sec:setup}

\paragraph{Benchmarks.} 
This work uses LongMemEval~\cite{wu2025longmemevalbenchmarkingchatassistants}, a recall benchmark on long user-assistant dialogues whose questions ask about content the user stated many turns earlier. 
The evaluation is restricted to the single-session-user subset, where each question is paired to the one earlier turn that holds its answer.
From each retained question the answer-bearing user turn is taken as the fact to store, and numeric-answer questions are removed which leaves 33 non-numeric facts. 
$N$ of these turns are written into the parameters and queried by the original held-out question, and recall is reported as $N$ grows from $4$ to $32$. 
The single-session subset supplies only 33 such facts, and $N{=}32$ therefore stores nearly the entire available pool, which places the study at the benchmark's ceiling rather than at a truncated sample. 
Recall is measured at every step from $N{=}4$ to $N{=}32$.
This shows how much the stored turns interfere as $N$ increases.
A generation counts as correct when the normalized ground-truth answer is contained in it and when a bidirectional DeBERTa-MNLI gate~\cite{he2020deberta,williams2017broad} judges it to entail the ground-truth answer.

\begin{table}[t]
\centering
\small
\setlength{\tabcolsep}{5pt}
\begin{tabular}{lccclc}
\toprule
\textbf{Model} & $\ell$ & $f$ & $\lambda$ & \textbf{Decode} & Self-Q \\
\midrule
Qwen3 (14B)  & 34 & 1 & 20 & beam-4, 4-gram block & $1,1$ \\
Nemo (12B)   & 31 & 1 & 40 & beam-4, 4-gram block & $5,3$ \\
Yi-1.5 (9B)  & 44 & 1 & 80 & greedy & $3,3$ \\
Llama-3 (8B) & 28 & 3 & 20 & beam-4, 4-gram block & $1,1$ \\
Mistral (7B) & 22 & 1 & 20 & greedy, rep.\ 1.3 & $1,1$ \\
\bottomrule
\end{tabular}
\caption{
Configuration for each language model.
Here, $\ell$ is the edited feed-forward layer, $f$ the first-token write strength, $\lambda$ the ridge coefficient, and Self-Q the anchored-question count $k_q$ and kept margin count $m$.
Decode is a 4-beam search blocking repeated 4-grams or greedy decoding.
}
\label{tab:configs}
\end{table}

\paragraph{Models.} 
This work uses five different models for evaluation; Qwen3 (14B)~\cite{qwen3technicalreport}, Mistral-Nemo (12B)~\cite{mistral2024nemo}, Yi-1.5 (9B)~\cite{01ai2024yi15}, Llama-3 (8B)~\cite{llama3modelcard}, and Mistral (7B)~\cite{mistral2024instructv03}.
The five come from four independently developed families, and recall on all of them tests whether the mechanism is general rather than tuned to one model.
All are widely used open-weight instruction-tuned models, which matches the dialogue setting the method targets and keeps the evaluation reproducible.
Each model's configuration is listed in Table~\ref{tab:configs}. 
Every run uses one A100 80GB GPU.
The forward pass runs in BF16 and each closed-form solve is carried out in FP32.

\subsection{Recall Attribution}
\label{sec:main}
Figure~\ref{fig:headline} compares \textsc{Skim} to four reference conditions at $N{=}32$ for five different models.
The `KV cache' shows recall when all models use the KV cache. 
`True-Q' shows recall without the KV cache when the true answer and the true question are given. 
`Self-Q' shows recall the same as True-Q without the KV cache and giving the true answer, but using the questions generated by the model. 
The `no-cache' shows recall when all models do not use the KV cache.
Here, the models that do not use the KV cache (`no-cache') show recall of 3.1 to 15.6\%, but this is difficult to regard as recall from the past dialogue because the model answers some questions by prior knowledge or by chance even without the turn.
The KV cache and the True-Q condition show recall of about 90 to 100\%, Self-Q shows recall of 80\% or more, and \textsc{Skim} shows recall of about 46.9 to 75.0\%.
Because the value to store is given as the ground-truth answer, True-Q and Self-Q show that the parameters can store and address a memory as well as the external memory. 
When the true question is swapped for the self-questions at a fixed answer, recall drops by about 6.88\% on average but still averages about 91.3\%. 
This indicates that the model can store and recall correctly even when its self-generated questions are used as the keys.
Replacing the answer by self-extracted spans lowers recall by about 26.9\% on average for all models, but recall still averages about 63.1\%.

Figure~\ref{fig:n_scaling_recall} shows the recall of \textsc{Skim} for each model as $N$ grows from $4$ to $32$. 
Every model starts near 100\% at small $N$ and declines gradually. 
Qwen3 shows the highest recall (75.0\%) and Mistral the lowest (46.9\%) at $N{=}32$. 
Figure~\ref{fig:n_scaling_fail} shows the count of unrecalled facts for each model as $N$ increases from $4$ to $32$, and Qwen3 fails to recall 8 of 32 facts and Mistral 17 at $N{=}32$. 
Both the recall and the unrecalled-fact count change gradually and show no sudden step at any size.
This gradual degradation indicates that recall declines by a small increment as each additional turn is stored, rather than failing abruptly once $N$ exceeds a threshold.
Even at $N{=}32$, \textsc{Skim} recalls well above the no-memory floor (`no-cache') and still remains effective as more turns are stored.

\begin{figure}[t]
\centering
\includegraphics[width=\columnwidth]{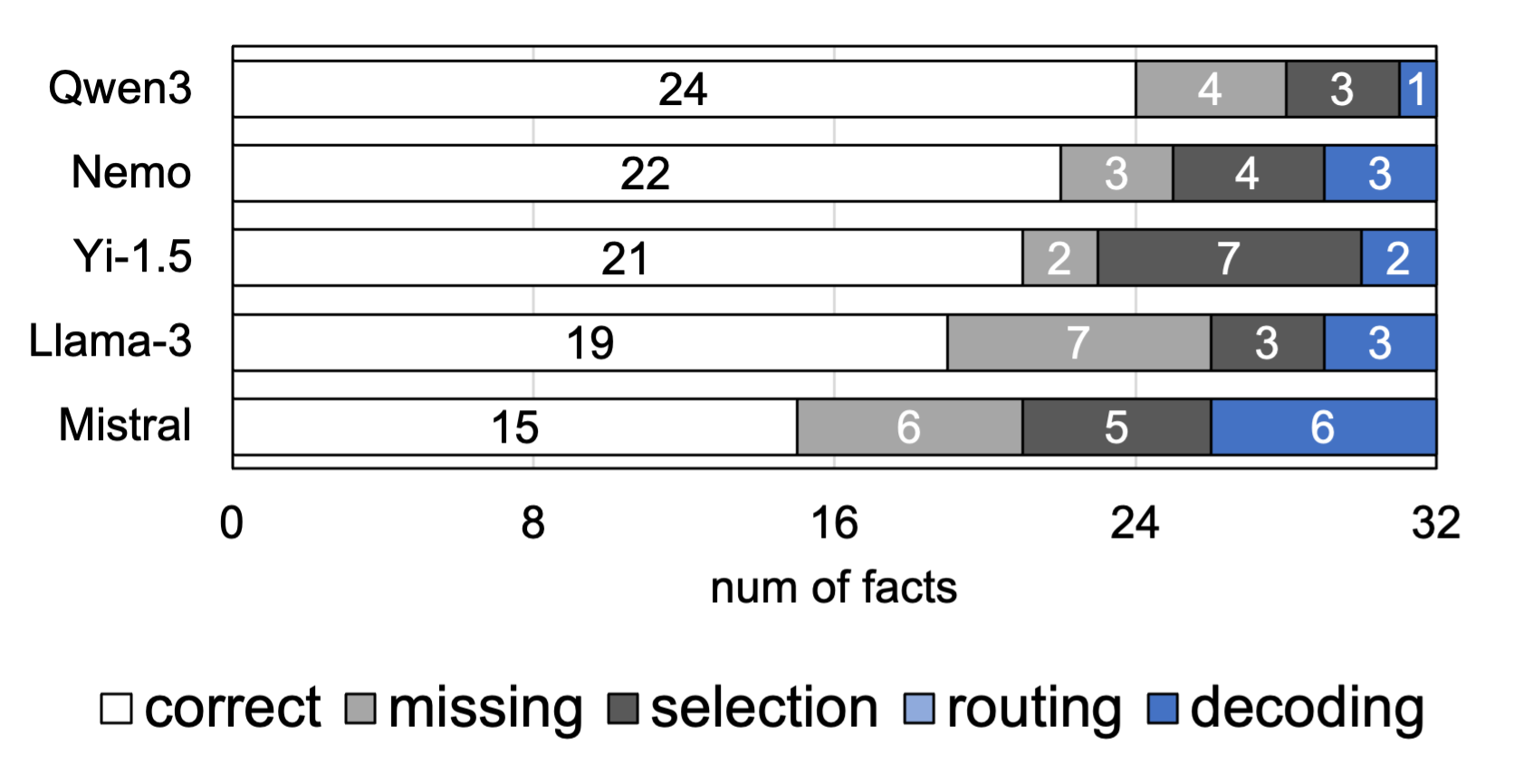}
\caption{
Per-fact outcome at $N{=}32$ for five language models. Each model's 32 stored facts split into correct and four failure classes, the store-time pair missing and selection and the recall-time pair routing and decoding.
}
\label{fig:53}
\end{figure}

\subsection{Which Stage Fails}
\label{sec:analysis}
For the five models, Figure~\ref{fig:53} shows the stage of \textsc{Skim} at which the 32 facts most often fail. 
A fact is `missing' when the true answer never enters the extracted span candidates, `selection' when a same-turn rival span is written in its place, `routing' when the user question retrieves a different fact, and `decoding' when the correct value is addressed but the generation breaks down. 
Here, `missing' and `selection' are store-time failures, and `routing' and `decoding' are recall-time failures.
The five models produce 59 failures; 22 from span-extraction (missing), 22 from selection, 0 from routing, and 15 from decoding.
Missing and selection account for 44 of the 59 failures, about three quarters of the total. 
Most failures therefore occur at store-time rather than at recall. 
Decoding accounts for 15 failures and forms the recall-time remainder. 
Routing is zero for every model, which indicates that the closed-form keys reach the intended fact and that read-time addressing is never where recall is lost.

The store-time failures split evenly overall, 22 missing and 22 selection in the five-model total.
Which mode dominates separates the models into two cases.
Selection dominates on Mistral-Nemo and Yi-1.5, which extract the correct span for 29 and 30 of the 32 facts and then rank a same-turn rival above it.
Yi-1.5 is the sharpest case at 2 missing and 7 selection.
Missing dominates on the other three models, which more often fail to extract the correct span at all.
Llama-3 is the sharpest case at 7 missing and 3 selection.
Thirteen types of intervention do not change the selection outcome and none corrects the selected value, whether they reshape the key geometry, enlarge the key budget, re-adapt the layer and regularizer, or widen the beam and align the stop criterion.
Because routing is zero and the keys already reach the intended fact, these interventions cannot correct a store-time failure.
Therefore, the limiting problem is the store-time extraction and selection of the span rather than any weakness in addressing or generation.

\begin{figure}[t]
\centering
\begin{subfigure}[t]{0.58\linewidth}
    \centering
    \includegraphics[width=\linewidth]{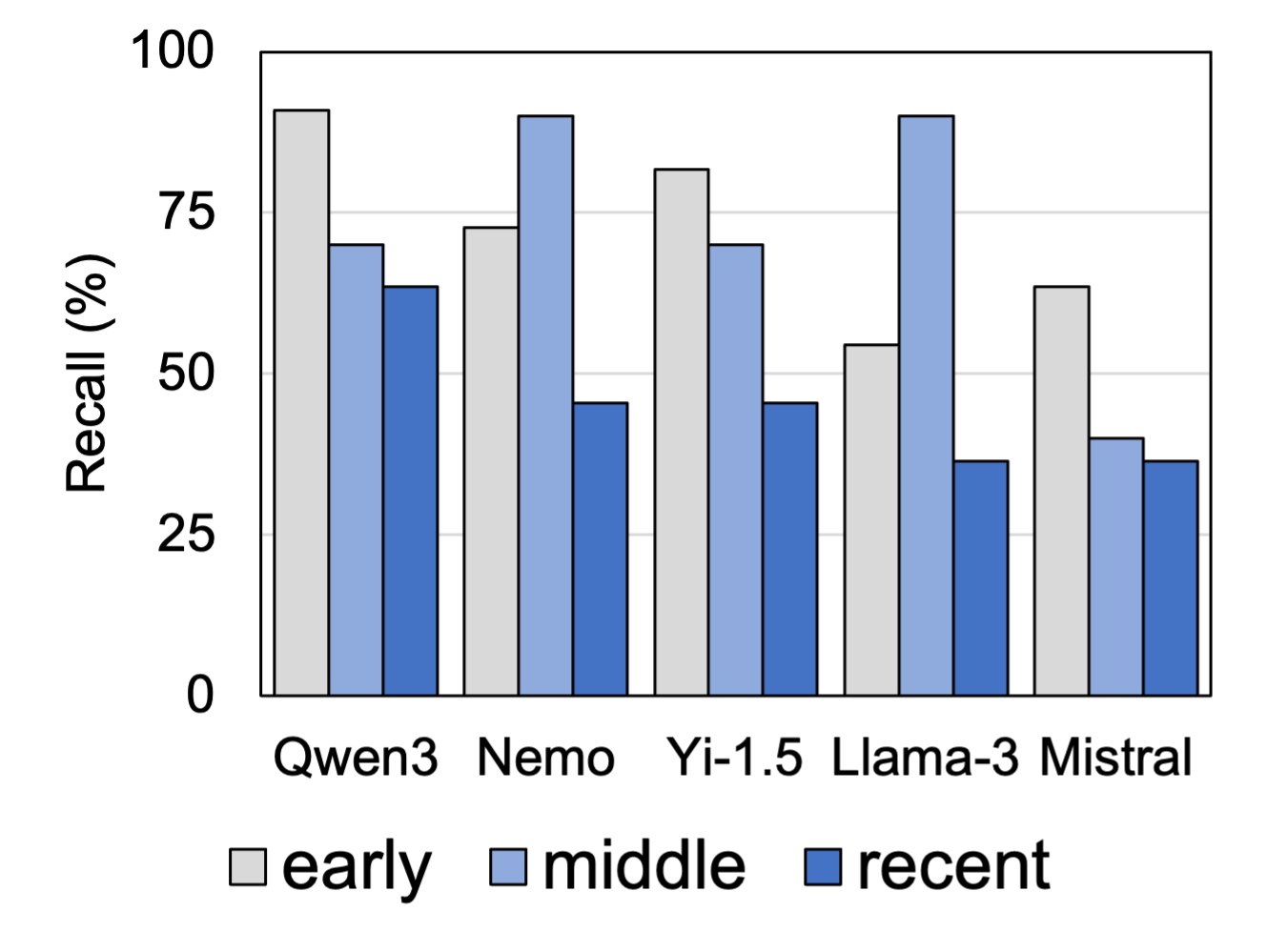}
    \caption{}
    \label{fig:54a}
\end{subfigure}
\hfill
\begin{subfigure}[t]{0.41\linewidth}
    \centering
    \includegraphics[width=\linewidth]{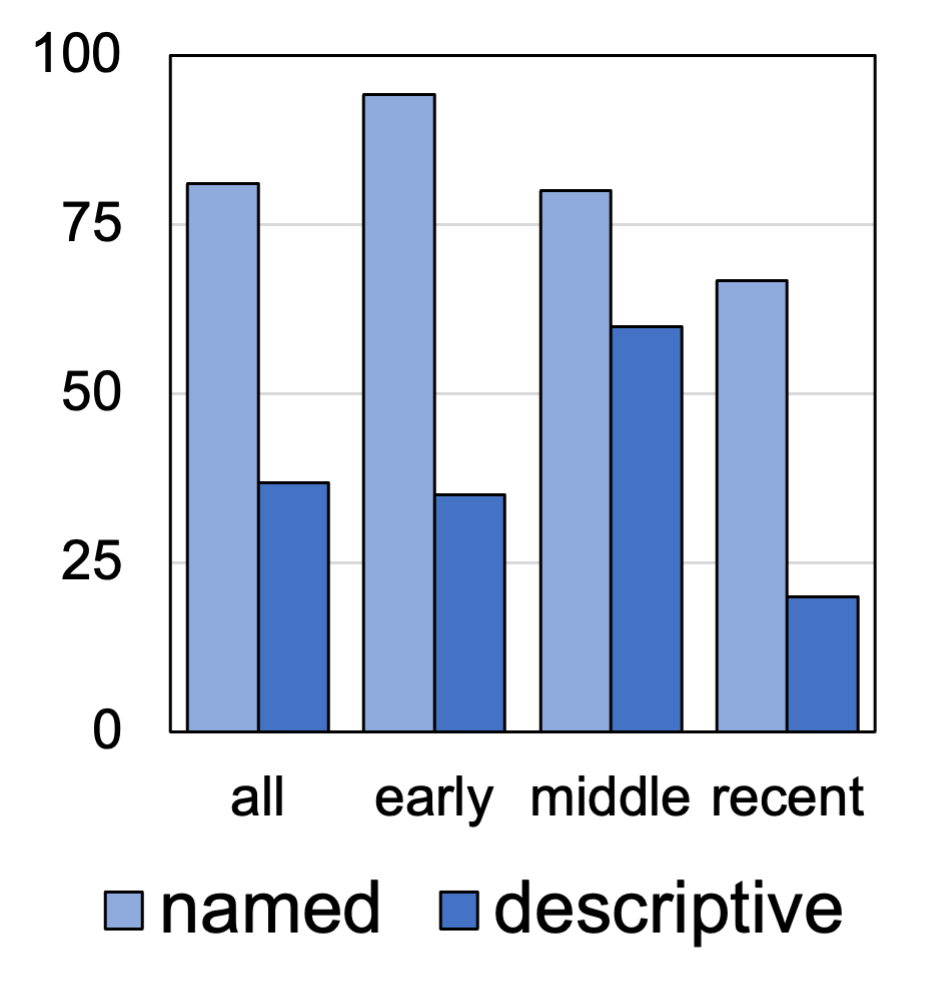}
    \caption{}
    \label{fig:54b}
\end{subfigure}
\caption{
Panel (a) reports recall by write position, each model's facts split into early, middle, and recent groups. Panel (b) reports recall by answer type, proper-noun versus descriptive multi-word, pooled and within each group.
}
\label{fig:54}
\end{figure}

\begin{figure}[t]
\centering
\includegraphics[width=\columnwidth]{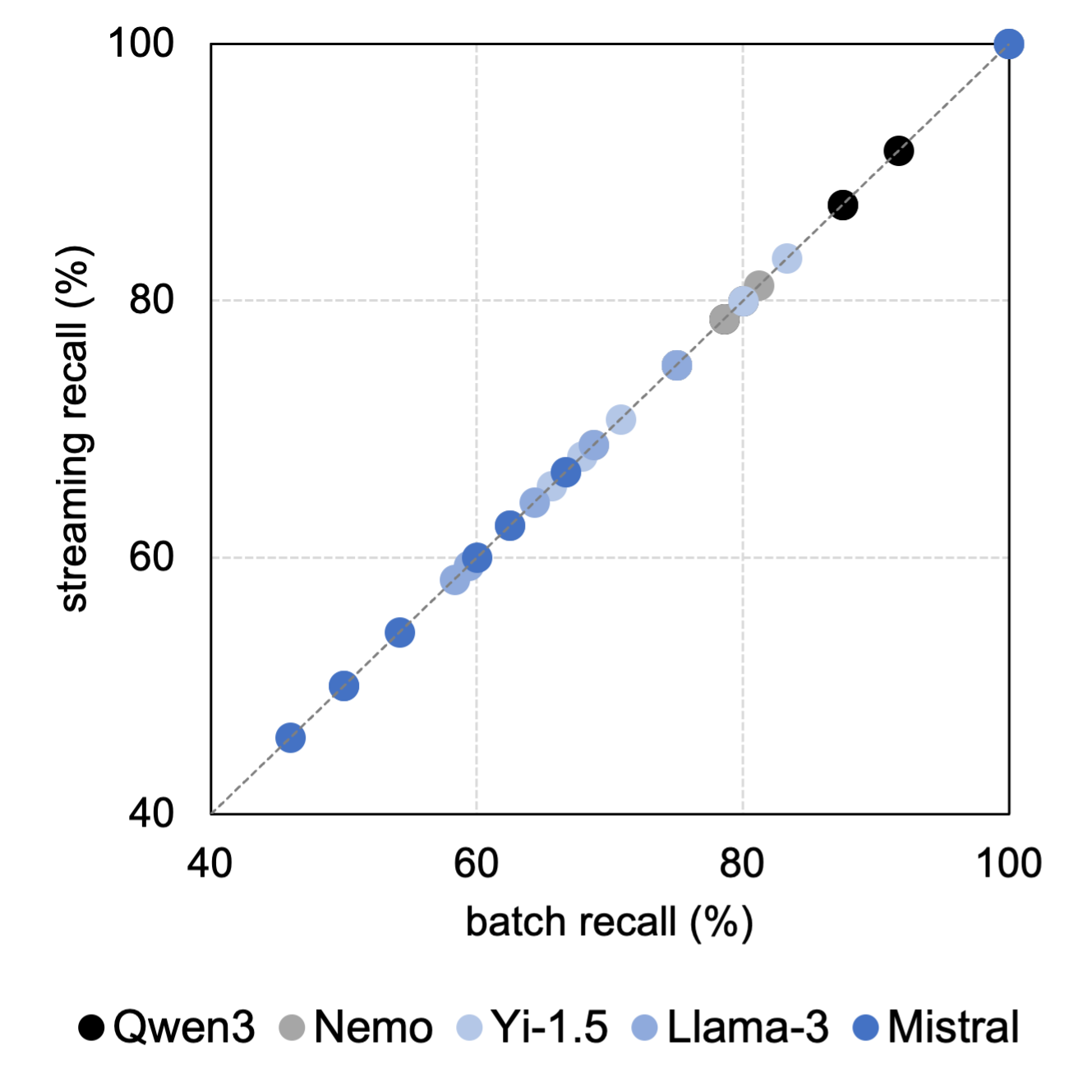}
\caption{
A comparison of streaming recall and batch recall for five models, ranging from $N{=}4$ to $N{=}32$. 
Every point lies on the identity line because a streamed write reproduces the batch solve without forgetting.
}
\label{fig:55}
\end{figure}

\subsection{Failure by Position and Answer Type}
\label{sec:whichfacts}
Figure~\ref{fig:54} reports the same $N{=}32$ outcomes two ways to separate two candidate axes of failure, the write position of a fact and the linguistic type of its answer.
Figure~\ref{fig:54a} groups the 32 facts by write position into an early group of 11, a middle group of 10, and a recent group of 11, and measures recall for each model in each group. 
In all five models, recall on the recent group is lower than on the early and middle groups. 
This apparent reduction in recall does not necessarily indicate that recent turns are forgotten.
Every turn is written at the same strength $f$ described in Eq.~\ref{eq:value}, and the edit therefore treats every write position equally.
The lower recall on the recent group instead follows from the answer type carried by those turns, which varies from turn to turn.

Figure~\ref{fig:54b} reports recall by answer type, and the split is sharp.
Proper-noun answers are recalled at 81.1\% while descriptive multi-word answers such as "a lighter shade of gray" reach only 36.9\%, a difference of about 44 points that largely holds within every position group.
The recent group appears weaker in Figure~\ref{fig:54a} only because it happens to carry more descriptive facts, and the apparent position effect is really an answer-type effect.
A proper noun is a discrete high-salience unit that the model extracts and keys cleanly.
A descriptive span instead has soft boundaries, runs to several tokens, and sits beside same-turn rivals.
Its failure is therefore one of self-extraction and multi-token reproduction rather than of storage or routing.
The hardest facts combine a descriptive answer and a same-turn rival span. 
The question-generation stage exists for exactly this structure and concentrates its per-fact gains there by recovering "Golden Retriever" rather than the rival "Max".
The descriptive-answer weakness is thus the clearest lever for improving self-supervision. 
It corroborates the earlier finding that self-extraction sets the recall ceiling rather than storage or addressing.

\subsection{Streaming Matches Batch}
\label{sec:streaming}
Figure~\ref{fig:55} plots streaming recall versus batch recall for all five models from $N{=}4$ to $N{=}32$.
Every point lies on the identity line, and a turn-at-a-time write reaches the recall of a single batch write at every size.
The match is exact because the two writes produce the same edit.
Each streaming step discards the previous edit and re-solves Eq.~\eqref{eq:accum} from the frozen base weights on all pairs collected to that point, and this solve reads only the accumulated Gram matrices $K_{\le t}^{\top}K_{\le t}$ and $D_{\le t}K_{\le t}^{\top}$.
These are order-invariant sums, and a permutation of the arrival order permutes the summands but not the sum, which makes the streamed edit the same matrix as the batch edit.
The figure confirms in measurement the permutation identity that Eq.~\eqref{eq:accum} forces in derivation.

The property depends on re-solving from the base weights each chunk rather than stacking each edit onto the previous one, an additive scheme that allows earlier turns to drift as later ones are written~\cite{fang2025alphaedit}.
A streamed dialogue therefore incurs no forgetting, and an earlier turn is recalled no worse once later turns are written on top of it.
This is the position-invariance of Figure~\ref{fig:54}(a) read from the write side, and the two jointly show that arrival order changes neither the stored content nor the recalled content.
A growing cache secures this same no-forgetting property only by keeping every turn, whereas the parametric store secures it at a size that stays fixed as the dialogue lengthens.
The re-solve is the one cost that grows, and it runs offline once per chunk.

\subsection{Discussion}
The main difficulty is the storage of a turn rather than its recall.
Recall of a stored value is near-perfect once that value is written, and Figure~\ref{fig:53} shows zero routing failures.
The difficulty is extracting from a turn which content to store without the future question or the answer.
The value harvest scores each candidate span by a self-question and a specificity prior and keeps the top few.
This fails most on descriptive multi-token spans beside same-turn rivals.
The other difficulty is the key rather than the value.
\textsc{Skim} keys each span by questions it writes before the real question arrives.
A later question phrased in an unforeseen way leaves its key unformed and its value unrecalled.
A separate limitation is the numeric answer.
A date or a count carries little of the specificity the harvest rewards and is rarely chosen for storage.
A recovered number also seldom reproduces the exact digits the containment metric checks for.
Moreover, prior weight-editing methods operate in a different environment from the one studied here~\cite{meng2022locating,meng2022mass,fang2025alphaedit}.
Running one on a dialogue turn would supply the ground-truth answer and question that this environment does not provide, turning it into an answer-given method rather than a comparable baseline.
The answer-given conditions of Figure~\ref{fig:headline} already serve as that method (see `True-Q'), and they mark the recall the correct answer would reach.

\section{Related Work}
Prior training-free methods modify model behavior or stored knowledge without gradient-based retraining~\cite{meng2022locating,fang2025alphaedit}.
Closed-form weight editing directly modifies selected parameters to store factual associations~\cite{meng2022locating,meng2022mass}.
ROME rewrites one subject-indexed factual association using a rank-one update to a single feed forward layer~\cite{meng2022locating}. 
MEMIT extends ROME to many predefined factual associations by distributing the required updates to several feed forward layers, enabling the simultaneous editing of thousands of associations~\cite{meng2022mass}.
AlphaEdit constrains each edit to the null space of the preserved knowledge's keys, which does not affect that knowledge and reduces interference during sequential edits~\cite{fang2025alphaedit}.
These methods require predefined edit targets and desired factual outputs~\cite{meng2022locating,meng2022mass,fang2025alphaedit}.
The proposed method constructs both the stored value and its key from the turn itself, taking neither a predefined edit target nor a desired output, and recalls the selected facts in response to later queries phrased differently.

\section{Conclusion}
This paper studied whether a language model can internalize real dialogue turns into its parameters by one closed-form edit.
The study was conducted in an assumed environment that provides no ground-truth answer and no later question and no external memory, and in a scenario where the dialogue arrives one turn at a time.
The proposed self-keyed in-parameter memorization (\textsc{Skim}) extracts what to store from the turn and attaches self-generated questions as its keys.
\textsc{Skim} recalls a five-model mean of 63.1\% and 75.0\% on the strongest model at 32 stored turns, well above a no-cache floor of 3.1 to 15.6\%.
The method also shows that streaming the dialogue reaches the same recall performance as storing all turns at once, which means the forgetting caused by sequential storage can be mitigated.
These experimental results show that the proposed \textsc{Skim} stores conversation content in a fixed-size model parameter without any external memory.

\bibliography{7.references}

\end{document}